\documentclass[letterpaper, 10 pt, conference]{ieeeconf}
\IEEEoverridecommandlockouts                              
\renewcommand{\baselinestretch}{0.985}

\usepackage[T1]{fontenc}

\usepackage{cite}
\usepackage{color}
\let\labelindent\relax
\usepackage{amsmath,amssymb,amsfonts, enumitem}
\usepackage{bbm, dsfont}
\usepackage{algorithm, algpseudocode}
\usepackage{graphicx}
\usepackage{textcomp}
\usepackage{mathtools}
\usepackage{subcaption}

\newtheorem{lemma}{Lemma}
\newtheorem{theorem}{Theorem}

\newtheorem{assumption}{Assumption}

\newtheorem{proposition}{Proposition}

\newcommand{\bt}[1]{{\boldsymbol{b}_t^{\scalebox{.6}{$(#1)$}}}}
\newcommand{\ttat}[1]{\boldsymbol{\theta}_t^{\scalebox{.6}{$(#1)$}}}

\newcommand{\bbE}{\mathbb{E}}
\newcommand{\bbR}{\mathbb{R}}
\newcommand{\indic}{\mathds{1}}
\newcommand{\Tgd}{T_{\mathrm{GD}}}

\newcommand{\degg}{\mathrm{deg}_g}
\newcommand{\nn}{\nonumber}
\newcommand{\ccon}{N^{\mathrm{Con}}}
\newcommand{\cucon}{N^{\mathrm{Ucon}}} 
\newcommand{\cpcon}{N^{\mathrm{Con}}_{\mathrm{proj}}}
\newcommand{\cpucon}{N^{\mathrm{Ucon}}_{\mathrm{proj}}} 
\newcommand{\Ccon}{D^{\mathrm{Con}}}
\newcommand{\Cucon}{D^{\mathrm{Ucon}}} 
\newcommand{\Cpcon}{D^{\mathrm{Con}}_{\mathrm{proj}}}
\newcommand{\Cpucon}{D^{\mathrm{Ucon}}_{\mathrm{proj}}} 
\newcommand{\svdeq}{\overset{\mathrm{SVD}}=} 
\newcommand{\qreq}{\overset{\mathrm{QR}}=} 
\newcommand{\scalemath}[2]{\scalebox{#1}{\mbox{\ensuremath{\displaystyle #2}}}}
\newcommand{\SD}[2]{\mathrm{SD}_2(#1,#2)}
\newcommand{\AvgCons}{\textsc{Agree}_g}
\newcommand{\gradU}[1]{\mathrm{gradU}^{\scalebox{.6}{$(#1)$}}}

\newcommand{\Err}[1]{\mathrm{Err}^{\scalebox{.6}{$(#1)$}}}
\newcommand{\Errg}[1]{\mathrm{Err}_{g}^{\scalebox{.6}{$(#1)$}}}
 
\newcommand{\ConsErr}[2]{\mathrm{ConsErr}_{#1}^{\scalebox{.6}{$(#2)$}}}
\newcommand{\UconsErr}[2]{\mathrm{UconsErr}_{\scalebox{.6}{$#1$}}^{\scalebox{.6}{$(#2)$}}}

\newcommand{\econ}{\epsilon_{\mathrm{con}}}
\newcommand{\Tcon}{T_{\mathrm{con}}}
\newcommand{\Tpm}{T_{\mathrm{pm}}}
\newcommand{\Tcongd}{T_{\mathrm{con,GD}}}
\newcommand{\Tconinit}{T_{\mathrm{con,init}}}
\newcommand{\ninit}{n_{\mathrm{init}}}
\newcommand{\ngd}{n_{\mathrm{GD}}}
\newcommand{\Utilde}[2]{\boldsymbol{\widetilde{U}}_{#1}^{\scalebox{.6}{$(#2)$}}}
\newcommand{\Ubreve}[2]{\boldsymbol{\breve{U}}_{#1}^{\scalebox{.6}{$(#2)$}}}
\newcommand{\Uhat}[1]{\boldsymbol{\widehat{U}}^{\scalebox{.6}{$(#1)$}} }
\newcommand{\Ubar}[1]{\boldsymbol{\bar{U}}^{\scalebox{.6}{$(#1)$}} }
\newcommand{\Ustar}{{\boldsymbol{U}^\star}}
\newcommand{\Bstar}{{\boldsymbol{B}^\star}}
\newcommand{\Ttastar}{{\boldsymbol{\Theta}^\star}}

\newcommand{\Ttag}[1]{\boldsymbol{\Theta}_g^{\scalebox{.6}{$(#1)$}}}

\newcommand{\Ttat}[1]{\boldsymbol{\Theta}^{\scalebox{.6}{$(#1)$}}}
\newcommand{\delg}[1]{{\delta^{\scalebox{.6}{$(#1)$}} }}
\newcommand{\rhog}[1]{{\rho^{\scalebox{.6}{$(#1)$}} }}
\newcommand{\psig}[1]{{\psi^{\scalebox{.6}{$(#1)$}} }}

\newcommand{\W}{\boldsymbol{W}}
\newcommand{\U}[2]{\boldsymbol{U}_{#1}^{\scalebox{.6}{$(#2)$}}}
\newcommand{\Ug}{\boldsymbol{U}_{g}}
\newcommand{\Bg}{\boldsymbol{B}_{g}}
\newcommand{\Z}[2]{\boldsymbol{Z}_{#1}^{\scalebox{.6}{$(#2)$}}}

\newcommand{\B}[2]{\boldsymbol{B}_{#1}^{\scalebox{.6}{$(#2)$}}}
\newcommand{\X}[2]{\boldsymbol{X}_{#1}^{\scalebox{.6}{$(#2)$}}}
\newcommand{\R}[2]{\boldsymbol{R}_{#1}^{\scalebox{.6}{$(#2)$}}}
\newcommand{\Xt}{{\boldsymbol{X}_t}}
\newcommand{\yt}{\boldsymbol{y}_t}
\newcommand{\ytt}[1]{\boldsymbol{y}_t^{\scalebox{.6}{$(#1)$}}}
\newcommand{\alpg}[2]{{\alpha}_{#1}^{\scalebox{.6}{$(#2)$}}}

\newcommand{\y}{\boldsymbol{y}}
\newcommand{\graph}{\mathcal{G}}
\newcommand{\Sg}{\mathcal{S}_g}
\renewcommand{\S}{\mathcal{S}}
\newcommand{\Pperp}{\boldsymbol{{P_{U^{\star}}}_\perp}}
\newcommand{\sigmax}{{\sigma_{\textrm{max}}^\star}}
\newcommand{\sigmin}{{\sigma_{\textrm{min}}^\star}}

\newcommand{\ceta}{c_{\eta}}
\newcommand{\efin}{\epsilon}
\newcommand{\Df}[2]{\nabla f_{#1}^{\scalebox{.6}{$(#2)$}}}

\newcommand{\minsval}[1]{\sigma_{\textrm{min}}(#1)}

\newcommand{\maxsval}[1]{\sigma_{\textrm{max}}(#1)}

\newcommand{\qed}{$\hfill\blacksquare$}
\newcommand{\dspaceX}{\mathcal{X}}
\newcommand{\dspaceY}{\mathcal{Y}}
\newcommand{\Rspace}[1]{\mathbb{R}^{#1}}
\newcommand{\mat}[1]{\boldsymbol{#1}}
\newcommand{\eye}[1]{\boldsymbol{I}_{#1}}
\newcommand{\lr}[1]{\left(#1\right)}
\newcommand{\Ng}{\mathcal{N}_g(\graph)}
\newcommand{\Tta}{\boldsymbol{\Theta}}
\newcommand{\tta}{\boldsymbol{\theta}}
\def\BibTeX{{\rm B\kern-.05em{\sc i\kern-.025em b}\kern-.08em
    T\kern-.1667em\lower.7ex\hbox{E}\kern-.125emX}}
\begin{document}
\title{Diffusion-based Decentralized Federated Multi-Task Representation Learning}
\author{Donghwa Kang and Shana Moothedath~\IEEEmembership{Senior Member,~IEEE}
\thanks{D. Kang and S. Moothedath are with Electrical and Computer Engineering, Iowa State University. Email: \{dhkang, mshana\}@iastate.edu.}
 \thanks{This work is supported by NSF-CAREER 2440455 and NSF-ECCS 2213069.}}
\maketitle

\begin{abstract}
Representation learning is a widely adopted framework for learning in data-scarce environments to obtain a feature extractor or representation from various different yet related tasks.
Despite extensive research on representation learning, decentralized approaches remain relatively underexplored.
This work develops a decentralized projected gradient descent-based algorithm for multi-task representation learning.
We focus on the problem of multi-task linear regression in which multiple linear regression models share a common, low-dimensional linear representation.
We present an alternating projected gradient descent and minimization algorithm for recovering a low-rank feature matrix in a diffusion-based decentralized and federated fashion. 
 We obtain constructive, provable guarantees that provide a lower bound on the required sample complexity and an upper bound on the iteration complexity (i.e., the total number of iterations needed to achieve a certain error level, $\epsilon$) of our proposed algorithm. We analyze the time and communication complexity of our algorithm and show that it is fast and communication-efficient.
 We performed numerical simulations to validate the performance of our algorithm and compared it with benchmark algorithms.
\end{abstract}
\section{Introduction}
\label{sec:introduction}

Multi-task representation learning (MTRL) has emerged as a powerful machine learning paradigm for learning multiple related models by integrating data from diverse sources. 
Here, the relatedness of different models implies that the features of individual tasks lie in a common subspace. 
Learning this common subspace enhances both performance and sample efficiency, which underlies the success of large-scale models in natural language processing and computer vision \cite{radford2019language, radford2021learning, brown2020language, devlin2018bert}.
Most existing works on MTRL have focused on a centralized setting, where data is either stored at a single location or distributed but managed by a central server (fusion center) for aggregation \cite{du2020few, chen2022active, collins2021exploiting, thekumparampil2021statistically}.
Such reliance on a central server introduces communication bottlenecks, privacy concern, and single-point failures. 
Given the increasing computing capabilities of modern edge devices, there are substantial advantages in decentralizing the computational burden to local nodes.
This motivates decentralized federated learning frameworks, where each node processes local data and communicates only the subspace estimate with its neighbors, eliminating the need for a central coordinator.
Despite this promise, decentralized approaches remain underexplored in the context of MTRL. 

The principal challenge in decentralized MTRL lies in reaching consensus on a good subspace estimate (in the sense of subspace distance of an orthonormal basis) across the network while minimizing communication costs, which are typically far more expensive—in time, bandwidth, and energy—than local computation.
In our previous work \cite{moothedath2022fast}, we adapted the alternating (projected) gradient descent and minimization (AltGDmin) algorithm \cite{nayer2022fast} to a decentralized setting, showing that this framework is viable even in communication constrained and data-scarce regimes.

A key finding is that the choice of communication mechanism is crucial. 
Due to the non-convex nature of MTRL, conventional decentralized optimization methods such as decentralized gradient descent (DGD) \cite{nedic2009distributed} are not directly applicable to the MTRL setting, as highlighted in \cite{moothedath2022fast}.
Our prior approach instead employed a \emph{combine-then-adjust} strategy: nodes first reached consensus on their local gradients and then performed projected gradient descent updates.
This method effectively propagates local information and achieves theoretical guarantees of linear convergence in subspace distance with high probability. 
However, it requires multiple consensus rounds per iteration, which depends on the final accuracy $\epsilon$, resulting in substantial communication overhead.

To overcome this limitation, we draw inspiration from the \emph{diffusion} principle, a powerful aggregation mechanism originally developed in adaptive filtering, and it is also utilized in decentralized optimization, and in decentralized meta learning recently \cite{cattivelli2009diffusion, tu2012diffusion, kayaalp2022dif}. 
Diffusion follows an \emph{adapt-then-combine} scheme: nodes first perform local updates and then exchange their updated variables with neighbors.
This seemingly minor change yields major benefits: diffusion enables stable and robust learning dynamics, reduces communication per iteration, and preserves statistical efficiency. 
In this work, we extend this principle to the decentralized MTRL problems by integrating diffusion with the AltGDmin framework, leading to the proposed \emph{Dif-AltGDmin} algorithm.

The main contributions of this paper are threefold:
\begin{itemize}
    \item We develop \emph{Dif-AltGDmin}, a diffusion-based alternating GD and minimization approach to solve the decentralized multi-task representation learning problem. The proposed algorithm is \emph{federated} in nature: only subspace estimates are exchanged rather than raw data.
    \item We establish theoretical guarantees and prove the convergence and sample complexity bounds of the proposed Dif-AltGDmin algorithm. We show that the communication complexity is significantly reduced compared to the existing approach, Dec-AltGDmin \cite{moothedath2022fast}.
    \item We validated the effectiveness of the proposed algorithm over baselines via numerical simulations. In particular, we demonstrated that the proposed algorithm is effective even with a single aggregation step, and its execution time is comparable to that of the centralized counterpart.
\end{itemize}

\section{Notations and Problem Formulation}

\noindent\textbf{Notations.} We use bold uppercase letters for matrices, bold lowercase letters for vectors, and regular fonts for scalars. For a positive integer $K$,  $[K]$ denotes the set $\{1,...,K\}$. For a vector $\mat{x}$, $\|\mat{x}\|$ denotes the $\ell_2$-norm of $\mat{x}$, while for a matrix $\mat{M}$, $\|\mat{M}\|_F$ and $\|\mat{M}\|$ denote the Frobenius and spectral norms. 
For a tall matrix $\mat{M}$, $\mat{M}^\dagger:=(\mat{M}^\top\mat{M})^{-1}\mat{M}^\top$ is a pseudo-inverse of $\mat{M}$, and $\mat{P}_{\mat{M}}:=\mat{M}(\mat{M}^\top\mat{M})^{-1}\mat{M}^\top$ is an orthogonal projection onto the column space of $\mat{M}$.
For an orthonormal matrix $\mat{U}$, the projection operators onto the column space and its orthogonal complement are $\mat{P}_{\mat{U}}=\mat{U}\mat{U}^\top$ and $\mat{P}_{\mat{U}_\perp}=\mat{I}-\mat{U}\mat{U}^\top$. 
Given two orthonormal matrices $\mat{U}_1,\ \mat{U}_2\in\Rspace{d\times r}$, the subspace distance is
$\SD{\mat{U}_1}{\mat{U}_2}:=\|(\eye{}-\mat{U}_1\mat{U}_1^\top)\mat{U}_2\|$.

\noindent\textbf{Problem Setting.}
We consider $T$ tasks, where each task $t\in[T]$ is associated with i.i.d. samples $\{x_{ti},y_{ti}\}_{i=1}^n$ drawn from a distribution $\mu_t$ over the joint space $\dspaceX\times\dspaceY$. 
Here, $\dspaceX\subseteq \Rspace{d}$ and $\dspaceY\subseteq \Rspace{}$ denotes the input and output space. 
Each task follows a linear model
   $ y_{ti}=x_{ti}^\top\tta_t^\star,\quad\forall t=1,...,T\ {\rm and }\ i=1,...,n,$
where $\tta_t^\star\in\Rspace{d}$ is an unknown feature vector.
Stacking samples yields the compact form
\begin{align*}
    \yt = \Xt\tta^\star_t, \quad \Xt\in\Rspace{n\times d},\ \yt\in\Rspace{n}.
\end{align*}
Multi-task representation learning (MTRL) aims to collectively learn the common underlying representation of the tasks.
In this work, we focus on learning low-dimensional representations, where each feature vector $\tta_t^\star$ is assumed to lie in a common $r$-dimensional subspace of $\Rspace{d}$ with $r\ll\min\{d,T\}$, i.e., the feature matrix $\Tta^\star=[\tta_1^\star, \tta_2^\star, \cdots, \tta_T^\star]\in\Rspace{d\times T}$ has rank $r$.
Letting $\Tta^\star\svdeq\Ustar\mat{\Sigma}^\star \mat{V}^{\star\top}:=\Ustar \Bstar$ with orthonormal matrix $\Ustar \in \Rspace{d\times r}$ and $\Bstar:=\mat{\Sigma}^\star \mat{V}^{\star\top}=[\mat{b}_1^\star,...,\mat{b}_T^\star]$, $\Ttastar$ can be recovered by identifying the shared representation $\Ustar$ and $\mat{b}_t^\star$ for all $t=1,...,T$.
We denote the maximum and minimum singular values of $\mat{\Sigma}^\star$ by $\sigmax$ and $\sigmin$, and the condition number as $\kappa:=\sigmax/\sigmin$.
Note that we consider the high-dimensional, data-scarce setting ($n<d$) where the number of samples per task is significantly smaller than the feature dimension. Within this regime, we seek to reconstruct $\Tta^\star$ using as few samples $n$ as possible. 

\noindent\textbf{Decentralized Setting.}
The $T$ tasks are distributed across $L$ nodes. Node $g$ holds tasks indexed by $\Sg\subset [T]$, with $\S_1,...,\S_L$ partitions $[T]$, i.e., $\cup_{g=1}^L \Sg=[T]$ and $\S_g \cap \S_{j} =\emptyset$ if $g \neq j$, $g,j \in [L]$. 
The network topology is modeled by an undirected connected graph $\graph=(\mathcal{V},\mathcal{E})$, where $\mathcal{V}=[L]$ and $\mathcal{E} \subseteq \mathcal{V} \times \mathcal{V}$ denotes the set of edges.
Each node can only exchange information with its neighbors $\Ng:=\{j:(g,j)\in \mathcal{E}\}$. There is no central coordinating node, thus each node can recover a subset of feature vectors, only for tasks that are contained in $\Sg$, i.e., $\Tta^\star_g:=[\tta_t^\star,\ t\in\Sg]$.

\noindent\textbf{Learning Objective.}
The loss function for the $t^{\rm th}$ task is 
\begin{align*}
\ f_t(\mat{U},\mat{b}_t):=\bbE_{(x_{ti},y_{ti})\sim\mu_t}\left[(y_{ti}-x_{ti}^\top\mat{U}\mat{b}_t)^2\right].
\end{align*}
The decentralized MTRL problem (Dec-MTRL) is 
\begin{align}\label{eqn: minimization centralized}
    &\min_{\substack{\mat{U}\in \bbR^{d \times r}\\\mat{B}\in \bbR^{r \times T}}} f(\mat{U},\mat{B}):=\sum_{g=1}^{L}\sum_{t\in \Sg}\|\yt-\Xt\mat{U}\mat{b}_t\|^2,
\end{align}
where the shared representation $\mat{U}\in\Rspace{d\times r}$ is an orthonormal matrix whose columns span $\Ustar$ and the task-specific coefficients $\mat{B}=[\mat{b}_1, \ldots, \mat{b}_T]\in\Rspace{r\times T}$.
This factorization exploits the low-rank structure of $\Tta^\star$, reducing computational complexity by learning $dr+Tr$ parameters instead of $dT$, a substantial saving since $r \ll \min\{d, T\}$. 
In the decentralized setting, the tasks are distributed among the nodes, and each node $g \in [L]$ observes only a subset $\Sg\subseteq [T]$ of the tasks. 
    Communication between nodes is constrained within their neighboring nodes as defined by the network.
    Hence, node $g$ can only evaluate its local cost function
   $f_g(\Ug,\Bg)=\sum_{t\in\Sg}\|\yt-\Xt\Ug\mat{b}_t\|^2,$
    compute local estimates $\Bg$ and $\Ug$ based on $f_g(\Ug,\Bg)$,
    and must collaborate with its neighboring nodes to recover global $\mat{U}$.
\vspace{2 mm}

\noindent\textbf{Assumptions.}
Since no $\yt$ is a function of the whole matrix $\Tta^\star$, recovering $\Tta^\star$ from local task data requires an additional condition to ensure each local dataset contains a sufficiently rich representation of the common subspace.
The following assumption guarantees this property.
\begin{assumption}\label{assumption: right-incoherence}
    For each $t=1,...,T$, $\|\mat{b}_t\|^2\leqslant \mu^2 \frac{r}{T} \sigmax^2$ with a numerical constant $\mu>1$. The same bound holds for each $\mat{x}_t^\star$ since $\|\mat{x}_t^\star\|^2=\|\mat{b}_t^\star\|^2$.
\end{assumption}
This assumption originated in \cite{candes2008exact} and has been widely adopted for low-rank matrix learning \cite{nayer2022fast, lee2019neurips} and multi-task representation learning  \cite{lin2024fast,tripuraneni2021provable}.
Additionally, we employ the following two commonly used assumptions.
\begin{assumption}\label{assumption: iid}
    Each entry of $\Xt$ is independently and identically distributed (i.i.d.) standard Gaussian random variable.
\end{assumption}
Assumption \ref{assumption: iid} enables the use of probabilistic concentration bounds and is crucial for reliable subspace recovery in a data-scarce regime. 
It is commonly assumed in most works on MTRL with theoretical guarantees \cite{tripuraneni2021provable, collins2021exploiting, OurICML}, and potential
extensions beyond this assumption are part of our future work.
\begin{assumption}\label{assumption: connected graph}
    The graph, denoted by $\graph$ is undirected and connected, meaning any two nodes are connected by a path.
\end{assumption}
Assumption \ref{assumption: connected graph} ensures that local information can propagate, allowing all nodes to reach consensus. This is a standard assumption in decentralized control, estimation, and learning \cite{olshevsky2009convergence, olfati2004Consensus}. We do not assume a fully connected network.

\section{Diffusion-based Decentralized MTRL}\label{sec: algorithm}

This section presents the details of the proposed algorithm. We begin with the preliminary background and then introduce the proposed algorithm for solving Dec-MTRL.

\noindent{\bf Agreement Protocol.}\label{sssc: AvgCons}
To coordinate the estimation of the global/shared representation $\mat{U}$, nodes exchange information with their neighbors and update local variables via weighted averaging
(Line~\ref{step:cons}, Algorithm~\ref{alg:dec-altgd}).
This weighted-sum operation serves as a generic agreement mechanism.
While earlier works, such as \cite{moothedath2022fast, moothedath2022dec} employ this update in the form of average \textit{consensus}, alternative variants such as diffusion arise depending on which variables are exchanged between nodes; we will return to this distinction in the Dif-AltGDmin algorithm.
Let $\W\in\Rspace{L\times L}$ be a doubly stochastic weight matrix with $\W_{gj}=1/\degg$ if $j\in\Ng$ and $\W_{gj}=0$ otherwise, where $\degg=|\Ng|$. 
Define $\gamma(\W):=\max(|\lambda_2(\W)|,\ |\lambda_L(\W)|)$.
Under standard connectivity assumptions, repeating multiple rounds of agreement protocol yields an approximation of the global average within a desired error bound as given in the result below. 
\begin{proposition}\label{prop: avgcons}(\cite{olshevsky2009convergence})
    Consider the agreement algorithm in Algorithm \ref{alg:AvgCons} with doubly stochastic weight matrix $\W$. Let $z_\mathrm{true}:=\frac{1}{L}\sum_{g=1}^L z^{\scalebox{.6}{(in)}}_{g}$ be the true average of the initial values $z^{\scalebox{.6}{(in)}}_{g}$ across $L$ nodes. For any $\econ<1$, if the graph is connected and if $\Tcon\geqslant \frac{1}{\log(1/\gamma(\W))}\log(L/\econ)$, then the output after $\Tcon$ iterations satisfies
    $$\max_g |z^{\scalebox{.6}{(out)}}_{g}-z_\mathrm{true}|\leqslant \econ\max_g|z^{\scalebox{.6}{(in)}}_{g}-z_\mathrm{true}|.$$
    Proposition \ref{prop: avgcons} is stated for scalar consensus, but it naturally extends to matrix valued variables.
    Rewriting $\Tcon$ expression in terms of $\gamma(\W)$ and fixing $L$ and $\Tcon$ gives the connectivity requirement for desired accuracy 
\begin{align}\label{eqn: connectivity}
    \gamma(\W)\leqslant\exp\left(-C\frac{\log(L/\econ)}{\Tcon}\right).
\end{align}
This inequality highlights the trade-off between accuracy and communication: achieving higher accuracy necessitates stronger network connectivity, i.e., smaller $\gamma(\W)$.
\end{proposition}

\begin{algorithm}[t]
\caption{Agreement Algorithm (\textsc{Agree})}
\label{alg:AvgCons}
\begin{algorithmic}[1]
\State \textbf{Input:} $\Z{g}{\mathrm{in}},\ \forall g\in[L],\ \Tcon,\ \graph$
\State Initialize $\Z{g}{0} \leftarrow \Z{g}{\mathrm{in}}$
\For{$\tau = 0$ \textbf{to} $\Tcon-1$, for each $g\in[L]$}
        \State $\Z{g}{\tau+1} \leftarrow \Z{g}{\tau} + \sum_{j\in \Ng} \tfrac{1}{\deg_g}\left(\Z{j}{\tau}-\Z{g}{\tau}\right)$
\EndFor
\State \textbf{Output:} $\Z{g}{\mathrm{out}} \leftarrow \Z{g}{\Tcon}$
\end{algorithmic}
\end{algorithm}
\vspace{2 mm}

\noindent{\bf Decentralized Spectral Initialization.}\label{sssc: initialization}
Since the cost function in Eq.~\eqref{eqn: minimization centralized} is non-convex, a proper initialization is needed. 
We adopt the {\em decentralized truncated spectral initialization} first proposed in \cite{moothedath2022fast}.
The spectral initialization aims at computing an initial estimate $\U{g}{0}$ close enough to $\Ustar$, and the initialization process guarantees $\delg{0}$-accurate recovery of common representation matrix $\Ustar$ and $\rhog{0}$-accurate node-wise consistency as described in  Proposition \ref{prop: initialization}.
In addition, it also quantifies the required number of samples and iteration complexities.
The decentralized spectral initialization is summarized in Algorithm \ref{alg:dec-init} for completeness.
For additional details, please refer to Theorem 3.4 of \cite{moothedath2022fast} and its proof. 
In Proposition~\ref{prop: initialization}, we include an additional condition 3) that is not explicitly stated in \cite{moothedath2022fast} but a direct consequence.  

\begin{proposition}\label{prop: initialization}(\cite{moothedath2022fast}.)
    (Decentralized Spectral Initialization)     
    Consider the decentralized truncated spectral initialization algorithm and suppose that Assumptions \ref{assumption: right-incoherence}-\ref{assumption: connected graph} hold. Pick $\delg{0}$ and $\rhog{0}$.  
    If $\Tconinit\geqslant C\frac{1}{\log1/\gamma(\W)}(\log L+\log d+\log\kappa+\log(1/\delg{0})+\log(1/\rhog{0}))$, $\Tpm\geqslant C\kappa^2\log(d/\delg{0})$, and $nT\gtrsim\kappa^4\mu^2(d+T)\frac{r}{\delg{0}^2},$ 
    then w.p. at least $1-1/d$,
    \begin{enumerate}
        \item $\SD{\U{g}{0}}{\Ustar}\leqslant \delg{0}$
        \item $\max_{g\neq g'}\|\U{g}{0}-\U{g'}{0}\|_F\leqslant\rhog{0}$
        \item In addition, for some $\psig{0}\leqslant\rhog{0}$,\\
        $\max_{g\neq g'}\|\Pperp(\U{g}{0}-\U{g'}{0})\|_F\leqslant\psig{0}$.
    \end{enumerate}
    
   \begin{proof}
        The proofs of 1) and 2) can be found in Appendix C of \cite{moothedath2022fast}. For 3), note that $\|\Pperp\|=1$, hence for any $g\neq g'$, 
        \begin{align*}
            \|\Pperp(\U{g}{0}-\U{g'}{0})\|_F&\leqslant\|\Pperp\|\cdot\|\U{g}{0}-\U{g'}{0}\|_F\\
            &\leqslant\|\U{g}{0}-\U{g'}{0}\|_F,
        \end{align*}
        which directly establishes the claim.
   \end{proof}
\end{proposition}

\begin{algorithm}[t]
\caption{Decentralized Spectral Initialization}
\label{alg:dec-init}
\begin{algorithmic}[1]
\State \textbf{Input:} $\{\Xt,\yt\}_{t\in\Sg},\ \graph,\ \kappa,\mu,n,T,\Tpm,\Tconinit$
\State Let $\yt \equiv \ytt{00},\ \Xt \equiv \X{t}{00}$
\State $\alpg{g}{\rm in}\leftarrow 9\kappa^2\mu^2\frac{L}{nT}\sum_{t\in\Sg}\sum_{i=1}^n y_{ti}^2$
\State $\alpha_g \leftarrow \AvgCons(\alpg{g'}{\rm in},\ g'\in[L],\ \graph,\ \Tconinit)$
\State Let $\yt \equiv \ytt{0},\ \Xt \equiv \X{t}{0}$
\State $\y_{t,trnc}:=\yt\circ\indic_{\{y_{ti}^2\leqslant\alpha_g\}}$
\State $\Ttag{0}=\Big[\tfrac{1}{n}\Xt^\top \y_{t,trnc},\ t\in\Sg\Big]$
\State Generate $\Utilde{g}{\rm in}$ with i.i.d.\ standard Gaussian entries (same seed for all $g$)
\State $\Utilde{g}{\rm in}\qreq \U{g}{0}\R{g}{0}$, so $\U{g}{0}=\Utilde{g}{\rm in}(\R{g}{0})^{-1}$
\For{$\tau=1$ \textbf{to} $\Tpm$, each $g\in[L]$}  
    \State $\Utilde{g}{\rm in}\leftarrow \Ttag{0}{\Ttag{0}}^\top \U{g}{0}$
    \State $\Utilde{g}{0}\leftarrow \AvgCons(\Utilde{g'}{\rm in},\ \graph,\ \Tconinit)$
    \State $\Utilde{g}{0}\qreq \U{g}{0}\R{g}{0}$, and get $\U{g}{0}$
    \State $\U{g=1}{\rm in}=\U{g=1}{0},\quad \U{g}{\rm in}=\mathbf{0}$ for $g\neq 1$
    \State $\U{g}{0} \leftarrow \AvgCons(\U{g}{\rm in},\ \graph,\ \Tconinit)$ for $g\neq 1$
\EndFor
\State \textbf{Output:} $\U{g}{0}$
\end{algorithmic}
\end{algorithm}

\begin{algorithm}[t]
\caption{Diffusion-based Alternating Gradient Descent and Minimization (Dif-AltGDmin)}
\label{alg:dec-altgd}
\begin{algorithmic}[1]
\State \textbf{Input:} $\Xt,\ \yt,\ t\in\Sg,\ g\in[L],\ \graph$
\State \textbf{Output:} $\U{g}{\Tgd},\ \B{g}{\Tgd},\ \Ttag{\Tgd}=\U{g}{\Tgd}\B{g}{\Tgd}$
\State \textbf{Parameters:} $\eta,\ \Tcongd,\ \Tgd$
\State \textbf{Sample-split:} Partition $\Xt,\ \yt$ into $2\Tgd+2$ disjoint sets $\X{\tau}{\ell},\ \ytt{\ell},\ \ell=00,0,1,\dots,2\Tgd$
\State \textbf{Initialization:} Run Algorithm~\ref{alg:dec-init} to get $\U{g}{0}$
\For{$\tau=1$ \textbf{to} $\Tgd$, for each $g\in[L]$}
    \State Let $\yt\equiv\ytt{\tau},\ \Xt\equiv\X{\tau}{\tau}$
    \State $\bt{\tau}\leftarrow (\Xt\U{g}{\tau-1})^\dagger \yt \quad \forall t\in\Sg$
    \State $\ttat{\tau}\leftarrow \U{g}{\tau-1}\bt{\tau} \quad \forall t\in\Sg$
    \State Let $\yt\equiv\ytt{\tau+\Tgd},\ \Xt\equiv\X{\tau}{\tau+\Tgd}$
    \State $\Df{g}{\tau}\leftarrow \sum_{t\in\Sg}\Xt^\top(\Xt\U{g}{\tau-1}\bt{\tau}-\yt)\bt{\tau}^\top$
    \State {\bf Local update:} $\Ubreve{g}{\tau}\leftarrow \U{g}{\tau-1}-\eta L\Df{g}{\tau}$ \label{line:cons-U}
    \State {\bf Diffusion:} $\Utilde{g}{\tau}\leftarrow \AvgCons(\Ubreve{g}{\tau},\ \graph,\ \Tcongd)$ \label{step:cons}
    \State \textbf{Projection:} $\Utilde{g}{\tau}\qreq \U{g}{\tau}\R{g}{\tau}$, $\U{g}{\tau}\leftarrow \Utilde{g}{\tau}{\R{g}{\tau}}^{-1}$ \label{line:proj}
\EndFor
\State \textbf{Output:} $\U{g}{\Tgd},\ \B{g}{\Tgd}=[\bt{\Tgd},\ t\in[T]],\ \Ttag{\Tgd}=\U{g}{\Tgd}\B{g}{\Tgd}$
\end{algorithmic}
\end{algorithm}
\vspace{2 mm}

\noindent{\bf Diffusion-based Alternating Projected GD and Minimization (Dif-AltGDmin).}\label{ssc: Dif-AltGDmin}
We present the pseudocode in Algorithm~\ref{alg:dec-altgd}.
Starting from the prescribed initialization phase, at every round, the proposed Dif-AltGDmin algorithm executes the following two steps: (i) minimization step for $\Bg$, (ii) {diffusion-based projected gradient descent for $\Ug$.}

{\bf Step 1. $\mat{B}$ update:} Consider $g$-th node. At time step $\tau$, the node computes $\Bg$ by minimizing $f_g(\U{g}{\tau-1},\Bg)$,
using the latest estimate of the common representation matrix $\U{g}{\tau-1}$.
Since $\U{g}{\tau-1}$ is fixed, we solve column-wise least square 
\begin{align*}
\bt{\tau}=(\Xt\U{g}{\tau-1})^\dagger \yt,
\end{align*}
for all $t\in\Sg$ and obtain $\B{g}{\tau}=[\bt{\tau},\ t\in\Sg]$.

{\bf Step 2. $\mat{U}$ update:}
Once node $g$ obtains $\B{g}{\tau}$, the gradient descent step follows, where the local gradient is computed as 
$\Df{g}{\tau}:=\nabla_{\mat{U}}f_g(\U{g}{\tau-1},\B{g}{\tau})$
and obtain \textit{local update},
\begin{align}\label{eq:U_update}
\Ubreve{g}{\tau}\leftarrow \U{g}{\tau-1}-\eta L\Df{g}{\tau}.
\end{align}
Recall that the global cost function in Eq.~\eqref{eqn: minimization centralized} is the sum of $L$ local cost functions.
This motivates multiplying the local gradient by $L$ in Eq.~\eqref{eq:U_update} to approximate the true gradient of $f(\mat{U}, \mat{B})$, $\gradU{\tau}:=\sum_{g=1}^L\Df{g}{\tau}$.
Each node exchanges the local estimate with neighboring nodes via the agreement algorithm, \textsc{Agree} to get the \textit{averaged update}, $\Utilde{g}{\tau}$, and then locally performs QR decomposition $\Utilde{g}{\tau}\qreq\U{g}{\tau}\R{g}{\tau}$ to get the \textit{projected update}, $\U{g}{\tau}=\Utilde{g}{\tau}(\R{g}{\tau})^{-1}$.
 Note that each node updates task-specific parameter $\bt{\tau}$ locally and exchanges only $\Utilde{g}{\tau}$ with neighbors. Thus, the algorithm is inherently federated.
\textsc{Agree} outputs an approximated average of the inputs from all nodes. 
Thus, aggregating local update $\Ubreve{g}{\tau}$ via \textsc{Agree} approximates the average given by 
\begin{align*}
    \Uhat{\tau}&:=\frac{1}{L}\sum_{g=1}^L\lr{\U{g}{\tau-1}-\eta L\Df{g}{\tau}}=\Ubar{\tau-1}-\eta\gradU{\tau},
\end{align*}
where $\Ubar{\tau}:=\frac{1}{L}\sum_{g=1}^L\U{g}{\tau}$ is the true average of $\U{g}{\tau}$'s and $\gradU{\tau}$ is the true gradient.
Thus, the proposed decentralized update rule approximates the centralized gradient descent step (all parameter estimates available at a fusion center) gradient descent step $${ \U{}{\tau+1}=\U{}{\tau}-\eta \nabla_U(f(\U{}{\tau},\B{}{\tau})).}$$ 
The guarantee for Dif-AltGDmin is given in Theorem \ref{thm: main}.
\begin{theorem}\label{thm: main}
Suppose that Assumptions \ref{assumption: right-incoherence}-\ref{assumption: connected graph} hold. 
Consider the outputs $\U{g}{\Tgd}$ and $\Ttag{\Tgd}$ for all $g\in[L]$ of Algorithm \ref{alg:dec-altgd} initialized with Algorithm \ref{alg:dec-init}. Pick $\efin<1$ and let $\eta=0.4/n\sigmax^2$.
Assume that 
\begin{enumerate}[label=\alph*)]
    \item $\Tpm=C\kappa^2(\log d+\log \kappa)$ and\\
        $\Tconinit= C\frac{1}{\log(1/\gamma(\W))}(\log L+\log d+\log r+\log\kappa)$;
    \item $\Tgd=C\kappa^2\log(1/\efin)$ and\\
        $\Tcongd= C\frac{1}{\log(1/\gamma(\W))}(\log L+\log r+\log\kappa)$;
    \item $nT \geqslant C\kappa^6\mu^2(d+T)r(\kappa^2r+\log(1/\efin))$.
\end{enumerate}
Then, with probability at least $1-1/d$,
\begin{enumerate}
    \item The task parameter is recovered up to an $\efin$-error, i.e., $$\|\ttat{\Tgd}-\tta_t^\star\|\leqslant\efin\|\tta_t^\star\|;$$
    \item The subspace distance, for all $t\in \mathcal{S}_g$, $g\in[L]$, satisfies
    $$\SD{\U{g}{\Tgd}}{\Ustar}\leqslant\efin.$$
\end{enumerate}
\end{theorem}
\textbf{Time Complexity.}
The time complexity of the proposed Dif-AltGDmin algorithm is 
\begin{align}\label{eqn: time-complexity}
    \mathcal{\tau}_{\mathrm{time}}
    =(\Tconinit\cdot \Tpm)\varpi_{\mathrm{init}}+(\Tcongd\cdot\Tgd)\varpi_{\mathrm{gd}}
\end{align}
where $\varpi_{\mathrm{init}}$ and $\varpi_{\mathrm{gd}}$ denote the per-step computational costs for initialization and Dif-AltGDmin, respectively.
For Dif-AltGDmin, each iteration involves
(i) task-wise least-square updates with cost $O(ndr\cdot |\Sg|+nr^2\cdot|\Sg|)$; $\bt{\tau}=(\Xt\U{g}{\tau-1})^\dagger \yt$ requires $ndr\cdot |\Sg|$ for computing $\Xt\U{g}{\tau-1}$ and $nr^2\cdot|\Sg|$ for least-square,
(ii) the gradient evaluation with cost $O(ndr\cdot|\Sg|)$; $\Df{g}{\tau}=\sum_{t\in\Sg}\Xt^\top( \Xt\U{g}{\tau-1}\bt{\tau}-\yt)\bt{\tau}^\top$, and QR factorization with cost $O(dr^2)$ per node.
Since $\sum_{g=1}^L|\Sg|=T$, the aggregated computational cost over all $L$ nodes for Dif-AltGDmin step is $\varpi_{\mathrm{gd}}=O(ndrT)$.
Similarly, each PM round in the initialization phase requires $\varpi_{init}=ndrT$.
Substituting the above and parts a) and b) of Theorem \ref{thm: main} into Eq.~\eqref{eqn: time-complexity}, the total time complexity is
\begin{align*}
    \mathcal{\tau}_{\mathrm{time}}&=\underbrace{(\Tconinit\cdot \Tpm)\varpi_{\mathrm{init}}}_{\mathcal{\tau}_\mathrm{init}}+\underbrace{(\Tcongd\cdot\Tgd)\varpi_{\mathrm{gd}}}_{\mathcal{\tau}_\mathrm{gd}}\\
    &\approx C\kappa^2\frac{\max(\log^2d,\log^2\kappa,\log^2L,\log^2(1/\efin))}{\log(1/\gamma(\W))}\cdot ndrT.
\end{align*}
To make the improvement more transparent, we separate the time complexities for initialization and GD steps
\begin{align*}
    \mathcal{\tau}_\mathrm{init}
    &\approx \frac{C\kappa^2\max(\log^2d,\log^2\kappa,\log^2L)}{\log(1/\gamma(\W))} ndrT\\
    \mathcal{\tau}_\mathrm{gd}
    &\approx \frac{C\kappa^2\log(\frac{1}{\efin})\cdot \max(\log L,\log r,\log\kappa)}{\log(1/\gamma(\W))} ndrT.
\end{align*}
In contrast, the corresponding time complexities in \cite{moothedath2022fast} are
\begin{align*}
    \mathcal{\tau}_\mathrm{init}
    &\approx \frac{C\kappa^4\max(\log^2d,\log^2\kappa,\log^2L,\log^2(\frac{1}{\efin}))}{\log(1/\gamma(\W))} ndrT\\
    \mathcal{\tau}_\mathrm{gd}
    &\approx \frac{C\kappa^4\log(\frac{1}{\efin})\cdot \max(\log(\frac{1}{\efin}),\log L,\log d,\log\kappa)}{\log(1/\gamma(\W))} ndrT.
\end{align*}
Hence, our algorithm achieves:
\begin{enumerate}
    \item Improved scaling in $\kappa$: both complexities involve $\kappa^2$ rather than $\kappa^4$.
    \item Less dependence on $\log(1/\efin)$: we avoid $\log(1/\efin)$ in $\mathcal{\tau}_\mathrm{init}$ and reduce $\log^2(1/\efin)$ to $\log(1/\efin)$ in $\mathcal{\tau}_\mathrm{gd}$.
    This is because our $\Tcongd$ does not depend on $\log(1/\efin)$, which is a significant improvement over the approach in \cite{moothedath2022fast}.
    
    \item Less dependence on $d$: $\mathcal{\tau}_\mathrm{gd}$ does not involve $\log d$ compared to \cite{moothedath2022fast}.
\end{enumerate}
 We remark that, since all nodes operate in parallel, the actual elapsed time scales with the node with the heaviest load, i.e., $\mathcal{\tau}_\mathrm{gd}
    \approx C\kappa^2\frac{\log(1/\efin)\cdot \max(\log L,\log r,\log\kappa)}{\log(1/\gamma(\W))}\cdot ndr\max_g|\Sg|$.

\noindent\textbf{Communication complexity.}
At each communication round, each node exchanges a $d\times r$ matrix with its neighbors. 
Since node $g$ has degree $\deg_g$, the per-round communication cost is $O(dr\cdot \max_g\deg_g\cdot L)$.
Over all rounds of initialization and Dif-AltGDmin, the total communication complexity is
\begin{align*}
    &\mathcal{\tau}_{\mathrm{comm}}=(\Tconinit \Tpm+ \Tcongd \Tgd)\cdot \lr{drL(\max_g \deg_g)}\\
    &=drL(\max_g \deg_g)\frac{C\kappa^2\max(\log^2d,\log^2\kappa,\log^2L,\log^2(\frac{1}{\efin}))}{\log(1/\gamma(\W))}.
\end{align*}

\section{Proof of Theorem \ref{thm: main}}\label{sec: proof of main thm}
We first define the notation used and present some preliminary results that are relevant to our analysis.

\noindent{\bf Definitions:}
Consider local estimates $\U{g}{\tau}$ and local gradients $\Df{g}{\tau}$, for $g\in[L]$.
We define the average $\Ubar{\tau}$ and the gradient sum $\gradU{\tau}$ as  
\begin{align*}
    \Ubar{\tau}:=\frac{1}{L}\sum_{g=1}^L \U{g}{\tau}\quad\mathrm{and}\quad 
    \gradU{\tau}:=\sum_{g=1}^L\Df{g}{\tau}.
\end{align*}
Based on this, we define the true average $\Uhat{\tau}$ that the \textsc{Agree} algorithm aims to approximate and the \textit{consensus error} between the true average and the actual output $\Utilde{g}{\tau}$
\begin{align*}
    &\Uhat{\tau}:=\frac{1}{L}\sum_{g=1}^L(\U{g}{\tau-1}-\eta L\Df{g}{\tau})=\Ubar{\tau-1}-\eta \gradU{\tau}\\
    &\ConsErr{g}{\tau}:=\Utilde{g}{\tau}-\Uhat{\tau}.
\end{align*}

The inter-node consensus error, i.e., $U$ estimation error between nodes $g$ and $g'$, is defined as
\begin{align*}
    \UconsErr{g,g'}{\tau}:=\U{g}{\tau}-\U{g'}{\tau}.
\end{align*}
and define $\Pperp:=I-\Ustar\Ustar^\top$.
We also define the gradient deviation from its expected value
\begin{align*}
    \Err{\tau}:=\bbE[\gradU{\tau}]-\gradU{\tau}.
\end{align*}
Similarly, we define $\Errg{\tau}:=\bbE[\Df{g}{\tau}]-\Df{g}{\tau}$.

The following result provides guarantee of minimization step, which is proved in Appendix A of \cite{moothedath2022fast}.
\begin{proposition}\label{prop: min-B}
    (Min step for $\mat{B}$) Assume that, for all $g\in[L]$, $\SD{\U{g}{\tau-1}}{\Ustar}\leqslant\delg{\tau-1}$, for some $\delg{\tau-1}\geqslant 0$.
    If $n\geqslant\max(\log T,\log d, r)$, then, with probability at least $1-\exp(\log T+r-cn)$,
    \begin{enumerate}
        \item $\|\bt{\tau}\|\leqslant1.1\|\mat{b}_t^\star\|$
        \item $\|\ttat{\tau}-\tta_t^\star\|\leqslant1.4\delg{\tau-1}\|\mat{b}_t^\star\|$
        \item $\|\Ttat{\tau}-\Ttastar\|_F\leqslant1.4\sqrt{r}\delg{\tau-1}\sigmax$.
    \end{enumerate}
    If $\delg{\tau-1}\leqslant0.02/\sqrt{r}\kappa^2$ and $\max_{g\neq g'}\|\U{g}{\tau-1}-\U{g'}{\tau-1}\|_F\leqslant \rhog{\tau-1}$ with $\rhog{\tau-1}\leqslant c/\sqrt{r}\kappa^2$, then the above implies that
    \begin{enumerate}[label=\alph*)]
        \item $\minsval{\B{}{\tau}}\geqslant 0.9\sigmin$
        \item $\maxsval{\B{}{\tau}}\leqslant1.1\sigmax$.\hfill$\Box$
    \end{enumerate}
\end{proposition}

In the rest of this section, we provide the proof of Theorem \ref{thm: main}. 
Due to the space limitation, we omit the proof of our Lemmas.
Theorem \ref{thm: main} guarantees $\epsilon$-accurate recovery of task parameters as stated in part 1) and subspace distance stated in part 2), given that the iteration and sample requirements stated in parts a)-c) hold.
The proof proceeds following four main steps: in the first two steps, we prove parts 1) and 2) of Theorem \ref{thm: main} specifying the required conditions, followed by the next two steps explaining how the conditions in parts a)-c) are satisfied.

  \textbf{\em Step 1. Task parameter recovery:} 
    Consider subspace distance bound at $\tau=\Tgd$, i.e., $\SD{\U{g}{\Tgd}}{\Ustar}\leqslant\delg{\Tgd}$. If 
    \begin{align}
        &\delg{\Tgd}\leqslant\efin\quad \mathrm{and}\label{eqn: cond - efin}\\
        &n\gtrsim \max\lr{\log d,\log T, r},\nn
    \end{align}
     then part 1) of Theorem \ref{thm: main} directly follows by part 2) of Proposition~\ref{prop: min-B} since $\|\tta_t^\star\|^2=\|\mat{b}_t^\star\|^2$.

    \textbf{\em Step 2. Subspace distance decay:}
    We prove that the subspace distance decays up to $\epsilon$, by establishing the relation of subspace distance at time $\tau$ with respect to $\tau-1$.
    In this and the following steps, we assume that for all $g\in[L]$,
    \begin{itemize}
        \item $\SD{\U{g}{\tau-1}}{\Ustar})\leqslant\delg{\tau-1}$
        \item $\max_{g,g'}\|\UconsErr{g,g'}{\tau-1}\|_F\leqslant\rhog{\tau-1}$
        \item $\max_{g,g'}\|\Pperp\UconsErr{g,g'}{\tau-1}\|_F\leqslant\psig{\tau-1}$.
    \end{itemize}
    Lemma \ref{lem: Subspace distance} characterizes the subspace distance decay through a gradient descent and  projection step for $\mat{U}$.
    \begin{lemma}\label{lem: Subspace distance} Let $\eta=\ceta/n\sigmax^2$. Consider the $\tau$-th instance of Algorithm \ref{alg:dec-altgd}. If 
        \begin{align}
            &\delg{\tau-1}\leqslant\frac{0.02}{\sqrt{r}\kappa^2}\label{eqn: cond - delg - lem}\\
            &\rhog{\tau-1}\leqslant\frac{0.1}{\sqrt{r}\kappa^2}=:\cucon\label{eqn: cond - rhog - lem}\\
            &\psig{\tau-1}\leqslant\frac{0.1}{1.21\kappa^2}\delg{\tau-1}=:\cpucon\delg{\tau-1}\label{eqn: cond - psig - lem}\\
            &\|\ConsErr{g}{\tau}\|\leqslant0.01\frac{\ceta}{\kappa^2}=:\ccon\label{eqn: cond - ConsErr - lem}\\
            &\|\Pperp\ConsErr{g}{\tau}\|\leqslant0.01\frac{\ceta}{\kappa^2}\delg{\tau-1}=:\cpcon\delg{\tau-1}\label{eqn: cond - PconsErr - lem},
        \end{align}
        and if 
        \begin{align}\label{eqn: cond - samples - lem}
            nT\geqslant C\kappa^4\mu^2dr\quad\mathrm{and}\quad n\gtrsim \max(\log T, \log d, r),
        \end{align}
        then, with high probability (w.h.p),
        \begin{align}\label{eqn: SD - result - lemma}
         \SD{\U{g}{\tau}}{\Ustar}\leqslant(1-0.3\ceta/\kappa^2)\delg{\tau-1}=:\delg{\tau}
        \end{align}  
        for all $g\in[L]$.\hfill$\Box$
    \end{lemma}
    
    In the next step, we analyze how these requirements can be met.

\textbf{\em Step 3. Iteration and sample complexities of GD step:} 
\textbf{\em Step 3.1. Bounding consensus error, inter-node consensus error and their projections:} 
    The following lemma establishes the connection between the number of consensus iterations $\Tcongd$ and the consensus error $\|\ConsErr{g}{\tau}\|$, and its projection $\|\Pperp\ConsErr{g}{\tau}\|$.
    \begin{lemma}\label{lem: Conserr}
    (Bounding consensus error) 
    Suppose the conditions given in Eqs.~\eqref{eqn: cond - delg - lem}-\eqref{eqn: cond - psig - lem} and Eq.~\eqref{eqn: cond - samples - lem} hold. If
    \begin{align*}
        \Tcon\geqslant C\frac{1}{\log(1/\gamma(\W))}\log(L/\econ),
    \end{align*}
    then, w.h.p, we have the following bounds of consensus errors and their projections onto complement of $\Ustar$. 
    \begin{align}\label{eqn: ConsErr - bound}
            \|\ConsErr{g}{\tau}\|_F
            &{\leqslant\econ\Ccon}
    \end{align}
    \begin{align}\label{eqn: PConsErr - bound}
        \textrm{and\ }\|\Pperp\ConsErr{g}{\tau}\|_F
        &{\leqslant\econ\Cpcon\delg{\tau-1}},
    \end{align}
    {for all $g\in[L]$, where $\Ccon:=\sqrt{r}\frac{0.1\ceta+0.034\ceta L}{\kappa^2}$
     and $\Cpcon:=\sqrt{r}(\frac{0.1}{1.21\kappa^2}+1.4\ceta L)$.}\hfill$\Box$
\end{lemma}

Next, the following lemma analyzes the change that the bounds of inter-node consensus error and its projection undergo during update.
\begin{lemma}\label{lem: Bounding UconsErr} (Bounding inter-node consensus error)
    Suppose conditions Eqs.~\eqref{eqn: cond - delg - lem}-\eqref{eqn: cond - ConsErr - lem} and Eq.~\eqref{eqn: cond - samples - lem} holds. If
    \begin{align*}
        \Tcon\geqslant C\frac{1}{\log(1/\gamma(\W))}\log(L/\econ).
    \end{align*}
    Then, w.h.p., following bounds hold for inter-node consensus errors and their projections onto the complement of $\Ustar$. 
    \begin{align}
        \|\UconsErr{g,g'}{\tau}\|_F        &{\leqslant\econ\Cucon}=:\rhog{\tau}\label{eqn: uconserr - bound}\\
        \|\Pperp\UconsErr{g,g'}{\tau}\|_F&\leqslant\econ\Cpucon\delg{\tau-1}=:\psig{\tau} \label{eqn: Puconserr - bound}
    \end{align}
    {for all $g,g'\in[L]$, where $\Cucon:=\sqrt{r}\frac{0.31\ceta+0.11\ceta L}{\kappa^2}$ and $\Cpucon:=\sqrt{r}(\frac{0.31}{1.21\kappa^2}+0.4\frac{\ceta}{\kappa^2}+3.3\ceta L)$ }.\hfill$\Box$
\end{lemma}

\textbf{\em Step 3.2. Iteration and communication complexities in Dif-AltGDmin step:} 
    Under the conditions Eqs.~\eqref{eqn: cond - delg - lem}-\eqref{eqn: cond - PconsErr - lem}, Theorem~\ref{thm: main} guarantees that $\Tgd$ iterations of Dif-AltGDmin algorithm returns the subspace estimate whose subspace distance to $\Ustar$ is no more than $\efin$, i.e., $\delg{\Tgd}\leqslant\efin$ (Eq.~\eqref{eqn: cond - efin}). 
    Suppose the conditions are satisfied for all $\tau=1,...,\Tgd$.
    Based on the result of Lemma \ref{lem: Subspace distance} in Eq.~\eqref{eqn: SD - result - lemma},     
    we require that the subspace distance after $\Tgd$ of Dif-AltGDmin iterations should satisfy
    \begin{align*}
        \delg{\Tgd} = \big(1-0.3\frac{\ceta}{\kappa^2}\big)^{\Tgd}\delg{0} \leqslant \efin.
    \end{align*}
    Taking logarithms and rearranging expression for $\Tgd$, 
    \begin{align}
        &\log \delg{\Tgd} 
        = \Tgd \log\Big(1-0.3\frac{\ceta}{\kappa^2}\Big) + \log \delg{0} \leq \log \efin, \nonumber \\
        &\implies \Tgd \geq -\frac{\log(\delg{0}/\efin)}{\log\big(1-0.3\ceta/\kappa^2\big)}\label{eqn: Tgd_LB}.
    \end{align}
    Using the inequality 
    $
    -\frac{1}{\log(1-x)} = \int_0^x \frac{1}{1-t} dt \leq \frac{x}{1-x} \leq 2x \quad \text{for } x \leq \frac{1}{2},
    $
    and absorbing $\log \delg{0}$ into a constant $C$, we obtain a simple bound
    \begin{align*}
        \Tgd \geq C \frac{\kappa^2}{\ceta} \log(1/\efin),
    \end{align*}
    which also satisfies Eq.~\eqref{eqn: Tgd_LB}.    
    This establishes the required number of Dif-AltGDmin steps $\Tgd$, ensuring that Eq.~\eqref{eqn: cond - efin} holds.

    Next, we obtain the number of consensus iterations $\Tcongd$.
    Previous steps state that our analysis remains valid if the conditions Eqs.~\eqref{eqn: cond - rhog - lem}-\eqref{eqn: cond - PconsErr - lem} hold at each $\tau=1,...,\Tgd$. (Also, Eq.~\eqref{eqn: cond - delg - lem} should hold.)
    { From Step 3.1, we know that (inter-node) consensus errors and their projections are bounded as in Eqs.~\eqref{eqn: ConsErr - bound}-\eqref{eqn: Puconserr - bound}, if the conditions in Eqs.~\eqref{eqn: cond - rhog - lem}-\eqref{eqn: cond - PconsErr - lem} hold. 
    From Eqs.~\eqref{eqn: ConsErr - bound} and \eqref{eqn: cond - ConsErr - lem}, if we set $\econ\leqslant\frac{\ccon}{\Ccon}$ in round $\tau$, then Eq.~\eqref{eqn: cond - ConsErr - lem} holds in round $\tau$.
    Similarly, from Eqs.~\eqref{eqn: PConsErr - bound} and \eqref{eqn: cond - PconsErr - lem}, if we set $\econ\leqslant\frac{\cpcon}{\Cpcon}$ in round $\tau$, then Eq.~\eqref{eqn: cond - PconsErr - lem} holds in round $\tau$.
    Further, from Eqs.~\eqref{eqn: uconserr - bound} and \eqref{eqn: cond - rhog - lem}, if we set $\econ\leqslant\frac{\cucon}{\Cucon}$ in round $\tau$, then Eq.~\eqref{eqn: cond - rhog - lem} holds in round $\tau+1$. 
    Similarly, from Eqs.~\eqref{eqn: Puconserr - bound}, \eqref{eqn: cond - psig - lem}, and using~\eqref{eqn: SD - result - lemma}, if we set $\econ\leqslant\frac{\cpucon}{\Cpucon}(1-0.3\frac{\ceta}{\kappa^2})$ in round $\tau$, then Eq.~\eqref{eqn: cond - psig - lem} holds in round $\tau+1$.
    }
    Since these must hold simultaneously, we deduce that the consensus accuracy $\econ$ should satisfy
    \begin{align}\label{eqn: econ - min}
        \scalemath{0.9}{\econ\leqslant\min\lr{\frac{\ccon}{\Ccon}, \frac{\cpcon}{\Cpcon},\frac{\cucon}{\Cucon}, \frac{\cpucon(1-0.3\frac{\ceta}{\kappa^2})}{\Cpucon}}}.
    \end{align}
    
    Both Lemmas \ref{lem: Conserr} and \ref{lem: Bounding UconsErr} state that such an accuracy $\econ$ can be guaranteed by at least 
    \begin{align}\label{eqn: Tcongd - min}
        \Tcongd&\geqslant C\frac{1}{\log(1/\gamma(W))}\log(L/\econ)\nonumber\\
        &= C\frac{1}{\log(1/\gamma(W))}(\log L+\log(1/\econ))
    \end{align}
   consensus iterations.
    Rewriting Eq.~\eqref{eqn: econ - min} as
    \begin{align*}
        \scalemath{0.9}{\log\lr{\frac{1}{\econ}}
        \geqslant
        \log \max\lr{\frac{\Ccon}{\ccon}, \frac{\Cpcon}{\cpcon},\frac{\Cucon}{\cucon}, \frac{\Cpucon}{\cpucon(1-0.3\frac{\ceta}{\kappa^2})}}}
    \end{align*}
    and by substituting the explicit expressions of these constants, we require $\log(1/\econ)$ to be greater than all of the following
    \begin{itemize}
        \item $\log(\frac{\Ccon}{\ccon})
            \geqslant\log(\frac{\sqrt{r}(0.1\ceta/\kappa^2+0.034\ceta L/\kappa^2)}{0.01\ceta/\kappa^2})$;
        \item $\log(\frac{\Cpcon}{\cpcon})
            \geqslant\log(\frac{\sqrt{r}(0.1/1.21\kappa^2+1.4\ceta L)}{0.01\ceta/\kappa^2})$;
        \item $\log(\frac{\Cucon}{\cucon})
            \geqslant\log(\frac{\sqrt{r}(0.31\ceta/\kappa^2+5.1\ceta L\sqrt{r}\cdot0.02/\sqrt{r}\kappa^2}{0.1\ceta/\kappa^2})$;
        \item ${\log\lr{\frac{\Cpucon}{\cpucon(1-0.3\frac{\ceta}{\kappa^2})}}
            \geqslant\log\lr{\frac{\sqrt{r}(\frac{0.31}{1.21\kappa^2}+0.4\frac{\ceta}{\kappa^2}+3.3\ceta L)}{\frac{0.1}{1.21\kappa^2}(1-0.3\ceta/\kappa^2)}}}$.
    \end{itemize}
    Thus, we require
    \begin{align}\label{eqn: econ - order}
        \log(1/\econ)\gtrsim\log L+\log r+\log\kappa.
    \end{align}   
    Combining Eqs.~\eqref{eqn: Tcongd - min} and \eqref{eqn: econ - order}, the communication complexity
       \begin{align*}
        \Tcongd\geqslant C\frac{1}{\log(1/\gamma(W))}(\log L+\log r +\log\kappa).
    \end{align*}
    Note that setting $\Tcongd$ in round $\tau$ guarantees Eqs.~\eqref{eqn: cond - rhog - lem}, \eqref{eqn: cond - psig - lem} hold in $\tau+1$. 
    To ensure these conditions hold for all rounds, especially at $\tau=1$, we require proper initialization of  $\delg{0}$, $\rhog{0}$ and $\psig{0}$. This initialization is achieved by choosing $\Tpm$ and $\Tconinit$ as specified in Step 4.
    
    \begin{figure*}[ht]
    \begin{subfigure}{0.35\textwidth}
        \centering
        \includegraphics[width=\linewidth]{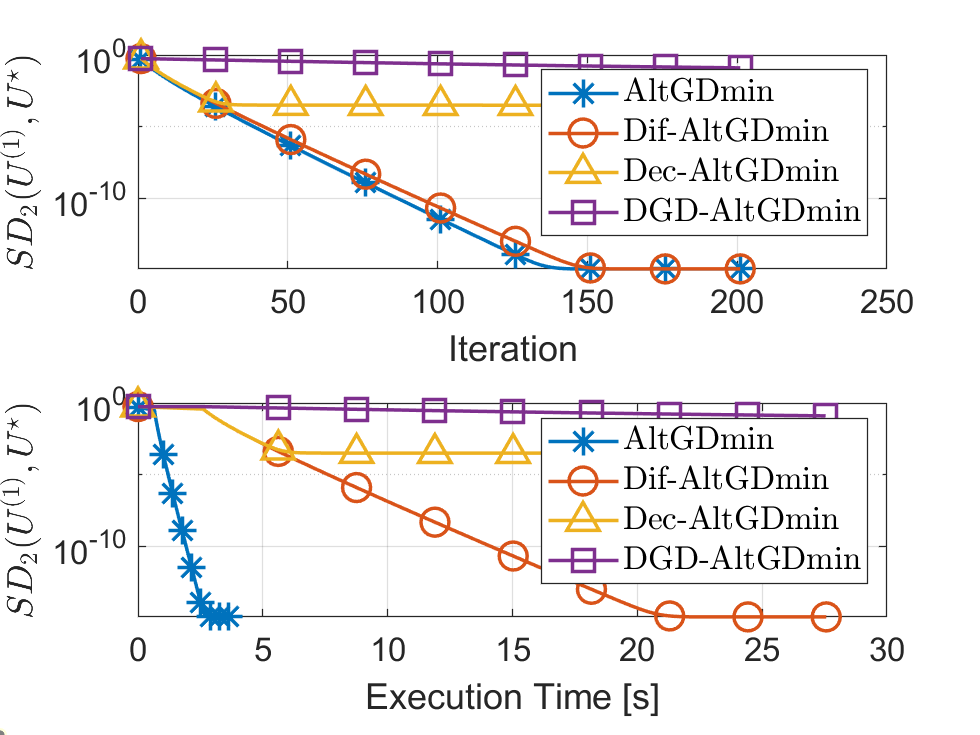}
        \vspace{-6 mm}
        \caption{$\Tcongd=\Tconinit=10$}
        \label{fig:1a}
    \end{subfigure}
    \hspace{-4 mm}
    \begin{subfigure}{0.35\textwidth}
        \centering
        \includegraphics[width=\linewidth]{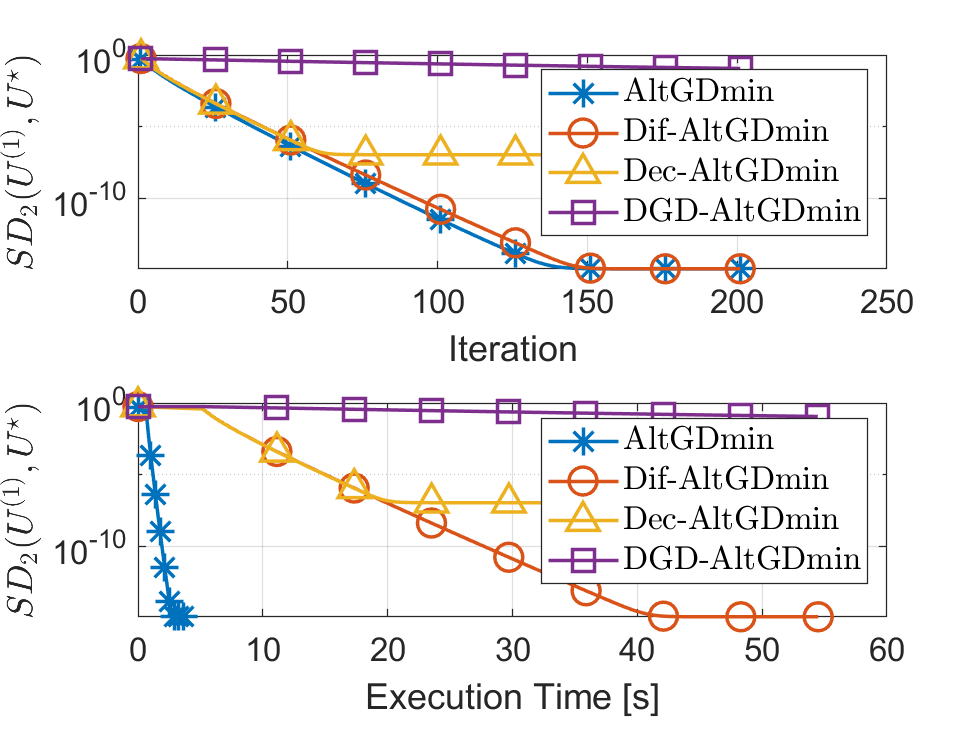}
        \vspace{-6 mm}
        \caption{$\Tcongd=\Tconinit=20$}
        \label{fig:1b}
    \end{subfigure}
     \hspace{-4 mm}
    \begin{subfigure}{0.35\textwidth}
        \centering
        \includegraphics[width=\linewidth]{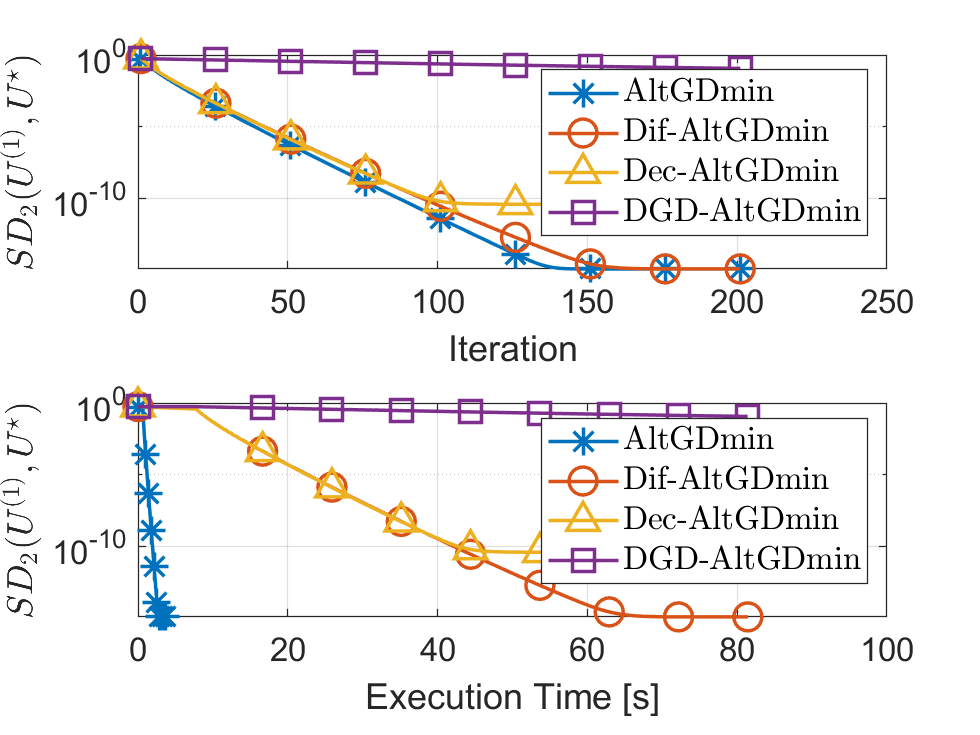}
        \vspace{-6 mm}
        \caption{$\Tcongd=\Tconinit=30$}
        \label{fig:1c}
    \end{subfigure}
    \vspace{-5 mm}
    \caption{Subspace distance vs. iteration count and execution time in seconds. We compare the performance of algorithms by varying number of consensus iterations, $\Tcon$. In all cases $\Tgd=500$, $L=20$, $d=T=600,\ r=4,\ n=30$, and $p=0.5$.
    }
    \label{fig:Varying Tcon}
\end{figure*}
    \textbf{\em Step 4: Iteration complexities of initialization step and total sample complexity:}
    {From part 3) of Proposition~\ref{prop: initialization}, we have $\psig{0}\leq \rhog{0}$, so it suffices to ensure
    }
    \begin{align*}
        \delg{0}\leqslant\frac{0.02}{\sqrt{r}\kappa^2}\quad
        \textrm{and}\quad\rhog{0}
        \leqslant\min(\frac{0.1\ceta}{\kappa^2}, \frac{0.1\delg{0}}{1.21\kappa^2})
        =\frac{c}{\sqrt{r}\kappa^4}.
    \end{align*}
    By Proposition \ref{prop: initialization}, these bounds determine the requirements for the initialization iterations:
    \begin{align}
        \Tpm&\geqslant C\kappa^2\log(d/\delg{0})=C\kappa^2(\log d+\log \kappa)\nn\\
        \Tconinit&\geqslant C\frac{1}{\log1/\gamma(\W)}(\log Ld\kappa+\log (\frac{1}{\delg{0}})+\log(\frac{1}{\rhog{0}}))\nn\\
        &= C\frac{1}{\log1/\gamma(\W)}(\log Ld\kappa r).\label{eqn: Tconinit}
    \end{align}
    
    Next, we bound the total number of samples required over both the initialization and Dif-AltGDmin phases.
    For initialization, we need
    $nT\gtrsim\kappa^8\mu^2(d+T)r^2=:\ninit T$, 
    and for each Dif-AltGDmin iteration,
    we need $n T\gtrsim C\kappa^4\mu^2dr=:\ngd T$.
    Therefore, the overall sample complexity is 
    \begin{align*}
        nT&\gtrsim \ninit T+\Tgd\cdot \ngd T\\
        &=\kappa^8\mu^2(d+T)r^2+C\kappa^6\mu^2dr\log(1/\efin)\\
        &=C\kappa^6\mu^2(d+T)r(\kappa^2r+\log(1/\efin)).
    \end{align*} 
    This completes the proof of Theorem \ref{thm: main}.\qed

\section{Simulations}

\begin{figure*}[ht]
    \begin{subfigure}{0.35\textwidth}
        \centering
        \includegraphics[width=\linewidth]{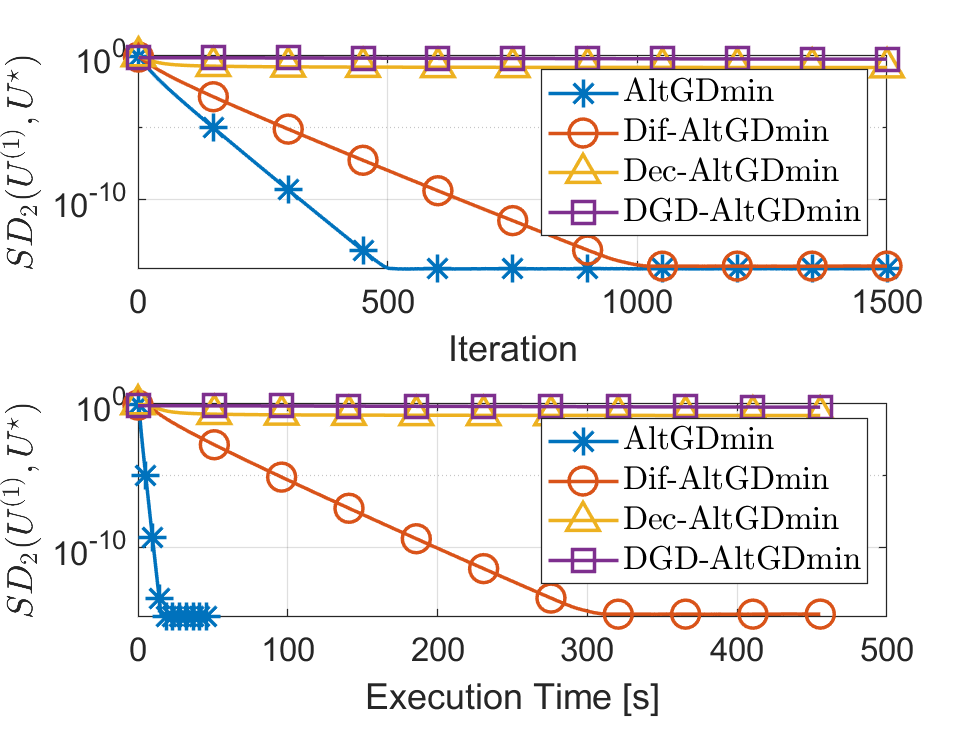}
        \vspace{-6 mm}
        \caption{$p=0.05$}
        \label{fig:2a}
    \end{subfigure}
        \hspace{-4 mm}
    \begin{subfigure}{0.35\textwidth}
        \centering
        \includegraphics[width=\linewidth]{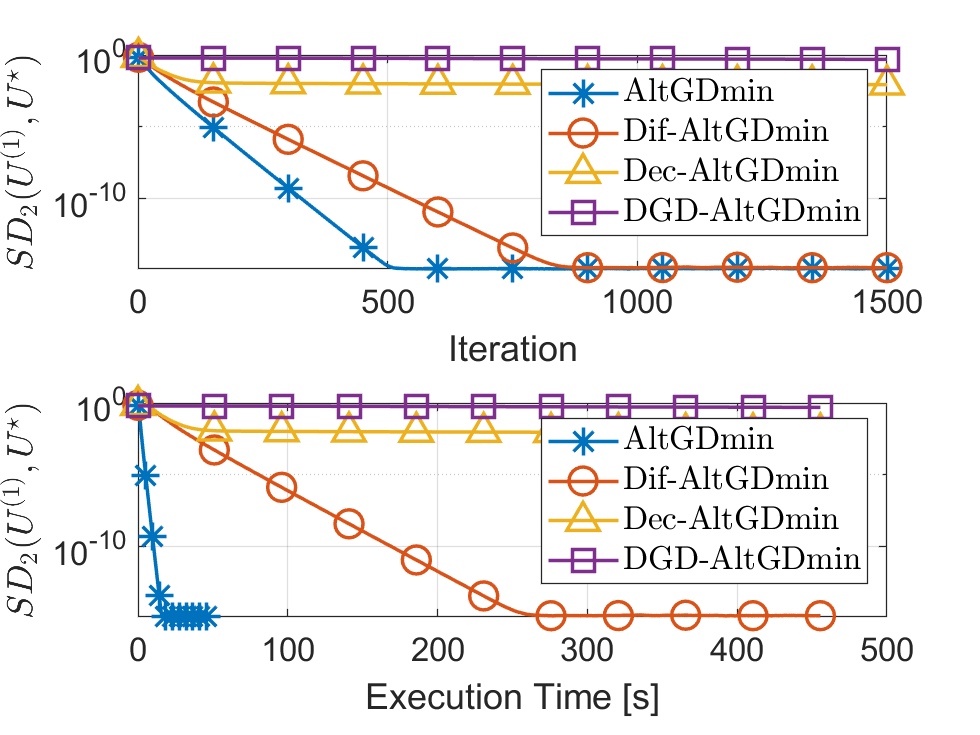}
        \vspace{-6 mm}
        \caption{$p=0.1$}
        \label{fig:2b}
    \end{subfigure}
        \hspace{-4 mm}
    \begin{subfigure}{0.35\textwidth}
        \centering
        \includegraphics[width=\linewidth]{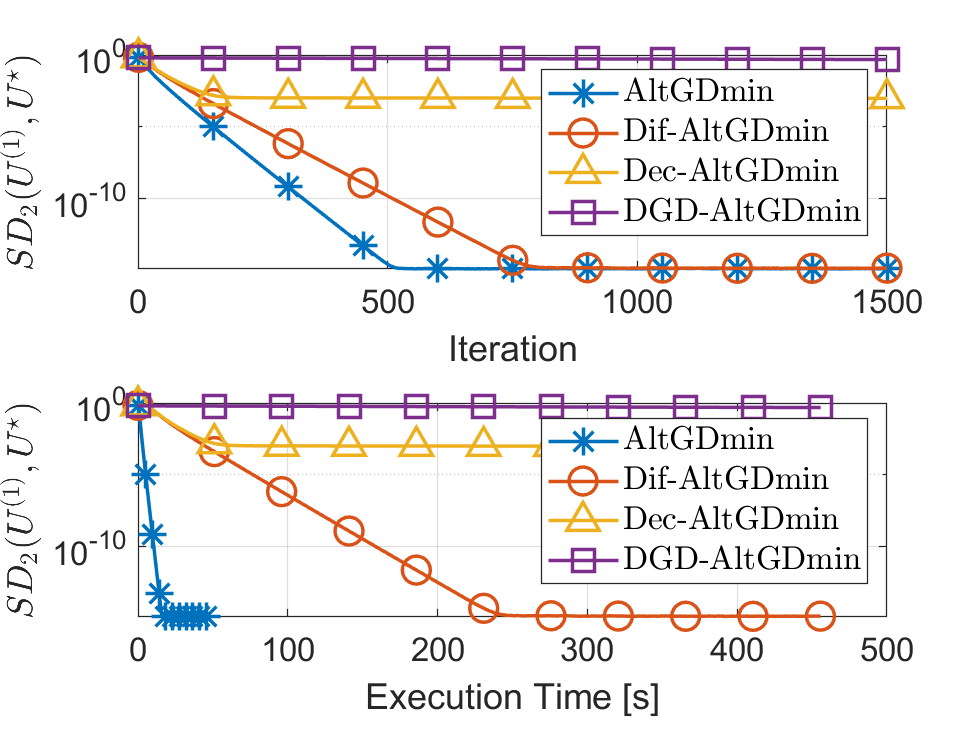}
        \vspace{-6 mm}
        \caption{$p=0.15$}
        \label{fig:2c}
    \end{subfigure}
    \vspace{-6 mm}
    \caption{Subspace distance vs. iteration count and execution time. We compare the performance of our algorithm by varying the edge probability in the communication graph, $p$. In all cases $\Tcongd=\Tconinit=10, \Tgd=1500$, $L=d=T=100, r=10$, and  $n=50$.}
    \label{fig: Varying p}
\end{figure*}

\noindent{\bf Data and network settings.}
All simulations are implemented in MATLAB using synthetic data. 
The communication network $\graph$ and the dataset $\{\Xt,\yt\}_{t=1}^T$ are randomly generated. 
Although our theoretical analysis relies on sample-splitting, we do not apply it in the simulations.
The learning rate is set to $\eta=\frac{0.4}{n\hat{\sigma}_{\rm max}^{\star^2}}$, where $\hat{\sigma}_{\rm max}^{\star^2}$ is the largest diagonal entry of $\R{g}{\Tpm}$.
The network is modeled as an Erdős-Rényi graph with $L$ nodes, where an edge between any two nodes exists independently with probability $p$.
To emulate the communication overheads, we assume a network bandwidth of $1\, \rm Gbps$ and a latency of $5\, m \rm s$.
Each transmitted entry is stored in double precision (8 bytes).
Therefore, at every \textsc{Agree} iteration ($\tau=1,\dots,\Tcongd$), the communication time per message is approximated as $t_{\rm comm}=50\times 10^{-3} + \frac{8dr}{10^9}+\tau_{\rm jitter}\;\;\rm seconds$, where the $\tau_{\rm jitter}$ is a small random perturbation to emulate network variability. 
We also assume that parallel communication capabilities: each node (or the central server in AltGDmin) can send and receive data from multiple neighbors simultaneously. Hence, only the maximum wall-clock time across all communications for a node is counted in the execution time.
For the evaluation, we plot the empirical average of the subspace distance $\SD{\U{1}{\tau}}{\Ustar}$ on the y-axis against the execution time or iteration count on the x-axis.
For a fair comparison, decentralized algorithms start from the same initial estimate $\U{g}{0}$ generated by Algorithm \ref{alg:dec-init}. All results are average of 100 independent trials.

\noindent{\bf Experiment 1.} In this experiment, we compare the performance of our Dif-AltGDmin algorithm with three baselines: (i) AltGDmin \cite{nayer2022fast}, a centralized algorithm that aggregates gradients from all $L$ nodes in one communication step and then broadcasts the updated $\mat{U}$; (ii) Dec-AltGDmin \cite{moothedath2022fast}, and {(iii) a DGD-variation of AltGDmin, defined as $\Utilde{g}{\tau}\leftarrow QR(\frac{1}{\degg}\sum_{g'\in\Ng} \U{g'}{\tau-1}-\eta\Df{g}{\tau})$.}
The experiment parameters are set as $L=20$, $d=600$, $T=600$, $n=30$, and $r=4$. 
To study the convergence and communication efficiency of the algorithms, we vary the values of $\Tcon$ as $10$, $20$ and $30$ (Figures \ref{fig:1a}-\ref{fig:1c}).
In contrast to decentralized algorithms that performs $\Tcongd$ agreement rounds each GD iteration, AltGDmin only communicates with nodes once for aggregation gradients and once for broadcasting $\mat{U}$ update. 
Also, due to the centralized nature, AltGDmin has full access to local gradients, while decentralized algorithms suffer consensus errors during updates.
Thus, AltGDmin achieves the fastest convergence speed.
Nevertheless, in all three cases, Dif-AltGDmin converges at the same order as AltGDmin with slightly more iterations. 
On the other hand, the subspace distance of Dec-AltGDmin decays almost identically with Dif-AltGDmin at earlier rounds but cannot reach below a certain bound that is heavily dependent on $\Tcongd$.
DGD-variant fails to converge effectively, as discussed in \cite{moothedath2022fast}.
The results demonstrate that Dif-AltGDmin is superior in conveterms of convergence and communication efficiency compared to other decentralized methods.

\noindent{\bf Experiment 2.} 
In this experiment, we evaluated the effect of network connectivity by  varying the edge probability $p$.
The results are presented in Figure \ref{fig: Varying p}.
We set the number of nodes equal to the number of tasks, with each node carrying a single task.
In all cases, Dif-AltGDmin achieves a comparable order of subspace distance to the centralized AltGDmin algorithm.
As the network connectivity improves, the number of iterations to reach that subspace distance level decreases slightly.
In contrast, the performance of Dec-AltGDmin is highly sensitive to network connectivity. 
Specifically, its performance improves with denser connectivity, yet remains significantly less accurate than that of AltGDmin and Dif-AltGDmin.
This behavior highlights the robustness of Dif-AltGDmin to weaker connectivity. 
Next, we provide an inference on why our algorithm is more robust to a sparse network than earlier work.
Recall the network connectivity expression in Eq.~\eqref{eqn: connectivity}. Consider $\Tcon :=\max{(\Tcongd,\Tconinit)}=\Tconinit$. 
While Dif-AltGDmin requires communication rounds as in Eq.~\eqref{eqn: Tconinit} 
to achieve $\efin$-accurate estimate,
Dec-AltGDmin requires a significantly tighter consensus accuracy $\log(1/\econ)\gtrsim \log (Ld\kappa\lr{1/\efin}^{\kappa^2})$
(Theorem 4.1 in \cite{moothedath2022fast}).
Since $\efin$ is typically very small, the consensus accuracy term $\log(1/\econ)$ for Dec-AltGDmin becomes substantially larger. 
Substituting $\log(1/\econ)$ expressions into Eq.~\eqref{eqn: connectivity}, we conclude that that Dec-AltGDmin imposes a much tighter upper bound on $\gamma(\W)$.
Smaller $\gamma(\W)$ indicates a denser network, meaning that Dec-AltGDmin requires a stronger network connectivity to guarantee its convergence.
Dif-AltGDmin, on the other hand, is robust to larger $\gamma(\W)$s and hence effective even under sparse networks.

\section{Conclusion}
We proposed \emph{Dif-AltGDmin}, a diffusion-based decentralized multi-task representation learning algorithm that combines local adaptation with efficient information exchange. 
Theoretical analysis established its convergence guarantees under the decentralized setting, and simulations confirmed its superiority over existing baselines in both accuracy and communication efficiency. 
Future work will focus on further reducing communication overhead by integrating techniques such as quantization, compression, and sporadic communication. 
In addition, we plan to validate the approach on real-world datasets using decentralized computing platforms such as Amazon Web Services (AWS).
\bibliographystyle{IEEEtran}
\bibliography{references.bib, bandits.bib}
\end{document}